\documentclass[10pt,twocolumn,letterpaper]{article}

\usepackage{cvpr}              

\usepackage{bbold}
\usepackage{newfloat}
\usepackage{listings}
\usepackage{multirow}
\usepackage{amssymb}
\usepackage{bm}
\usepackage{booktabs}
\usepackage{enumerate}
\usepackage{amssymb,mathrsfs,amsmath}
\usepackage{makecell} 
\usepackage{multirow}
\usepackage{times}
\usepackage{epsfig}
\usepackage{graphicx}
\usepackage{amsmath}
\usepackage{amssymb}
\usepackage{booktabs}
\usepackage{color}
\usepackage{mathtools}
\usepackage{mathtools}

\usepackage{amsmath}
\usepackage{amssymb}
\usepackage{xspace}
\usepackage{verbatim}
\usepackage{amsthm}

\theoremstyle{definition}

\theoremstyle{remark}

\makeatletter
\newcommand{\rmnum}[1]{\romannumeral #1}
\newcommand{\Rmnum}[1]{\expandafter\@slowromancap\romannumeral #1@}
\makeatother

\definecolor{cvprblue}{rgb}{0.21,0.49,0.74}
\usepackage[pagebackref,breaklinks,colorlinks,allcolors=cvprblue]{hyperref}

\def\paperID{1283} 
\def\confName{CVPR}
\def\confYear{2026}

\title{DICArt: Advancing Category-level Articulated Object Pose Estimation in Discrete State-Spaces}

\author{
Li Zhang\textsuperscript{ 1}\footnotemark[1], 
\quad Mingyu Mei\textsuperscript{ 3}\thanks{Equal Contribution.} ,
\quad Ailing Wang \textsuperscript{ 4},
\quad Xianhui Meng \textsuperscript{ 1},
\quad Yan Zhong \textsuperscript{ 5}, \\
\quad Xinyuan Song \textsuperscript{ 6},
\quad Liu Liu \textsuperscript{ 7, 8, 9} \thanks{Corresponding Author.},
\quad Rujing Wang \textsuperscript{ 1},
\quad Zaixing He\textsuperscript{ 3},
\quad Cewu Lu \textsuperscript{ 2} \\
\textsuperscript{1} University of Science and Technology of China \quad 
\textsuperscript{2} Shanghai Jiao Tong University \\
\textsuperscript{3} Zhejiang University \quad 
\textsuperscript{4} East China Normal University \quad 
\textsuperscript{5} Peking University \quad 
\textsuperscript{6} Emory University \\
\textsuperscript{7} Hefei University of Technology \quad 
\textsuperscript{8} Jianghuai Advance Technology Center \\ 
\textsuperscript{9} Anhui Provincial Key Laboratory of Humanoid Robots \\
\vspace{5pt}
{\tt zanly12138@gmail.com, mingyumei@zju.edu.cn}
}

\begin{document}
\maketitle
\begin{abstract}
Articulated object pose estimation is a core task in embodied AI. Existing methods typically regress poses in a continuous space, but often struggle with 1) navigating a large, complex search space and 2) failing to incorporate intrinsic kinematic constraints.
In this work, we introduce \textbf{DICArt} (\textbf{DI}s\textbf{C}rete Diffusion for \textbf{Art}iculation  Pose Estimation), a novel framework that formulates pose estimation as a conditional discrete diffusion process. 
Instead of operating in a continuous domain, DICArt progressively denoises a noisy pose representation through a learned reverse diffusion procedure to recover the GT pose.
To improve modeling fidelity, we propose a flexible flow decider that dynamically determines whether each token should be denoised or reset, effectively balancing the real and noise distributions during diffusion. Additionally, we incorporate a hierarchical kinematic coupling strategy, estimating the pose of each rigid part hierarchically to respect the object's kinematic structure.
We validate DICArt on both synthetic and real-world datasets. Experimental results demonstrate its superior performance and robustness. By integrating discrete generative modeling with structural priors, DICArt offers a new paradigm for reliable category-level 6D pose estimation in complex environments. Project website: \url{https://sites.google.com/view/dicartpub}.
\end{abstract}

\section{Introduction}\label{sec:intro}
Object pose estimation is a critical task with applications spanning robotics~\cite{billard2019trends, mason2001mechanics, shridhar2022cliport}, augmented reality~\cite{zhao2025zero, zhao2024wavelet, carmigniani2011augmented, amin2015comparative}, human-computer interaction~\cite{liu2021thin, preece1994human, myers1998brief}, and embodied AI~\cite{yu2023gamma,karrer2011pinstripe, hou2021exploring, jaritz2019multi}. Articulations, characterized by their inherent flexibility and joint structure, present unique challenges for pose estimation due to their non-rigid motion and complex interdependencies between the individual parts. Accurate 6D  \textbf{A}rticulation \textbf{P}ose \textbf{E}stimation (\textbf{APE task}) is essential for tasks like manipulation~\cite{katz2008manipulating, jiang2023vima}, tracking~\cite{xue2023garmenttracking, liu2024kpa}, and scene understanding~\cite{zhao2025ultrahr, yu2023comprehensive}. The ability to reliably estimate the poses of articulated objects facilitates more precise interaction with the environment, improving the performance of robotic systems and enhancing immersive experiences in virtual environments~\cite{zhao2024learning, manas2023robotic,zhao2024cycle}.

\begin{figure*}[t!]
    \centering
\includegraphics[width=0.88\linewidth]{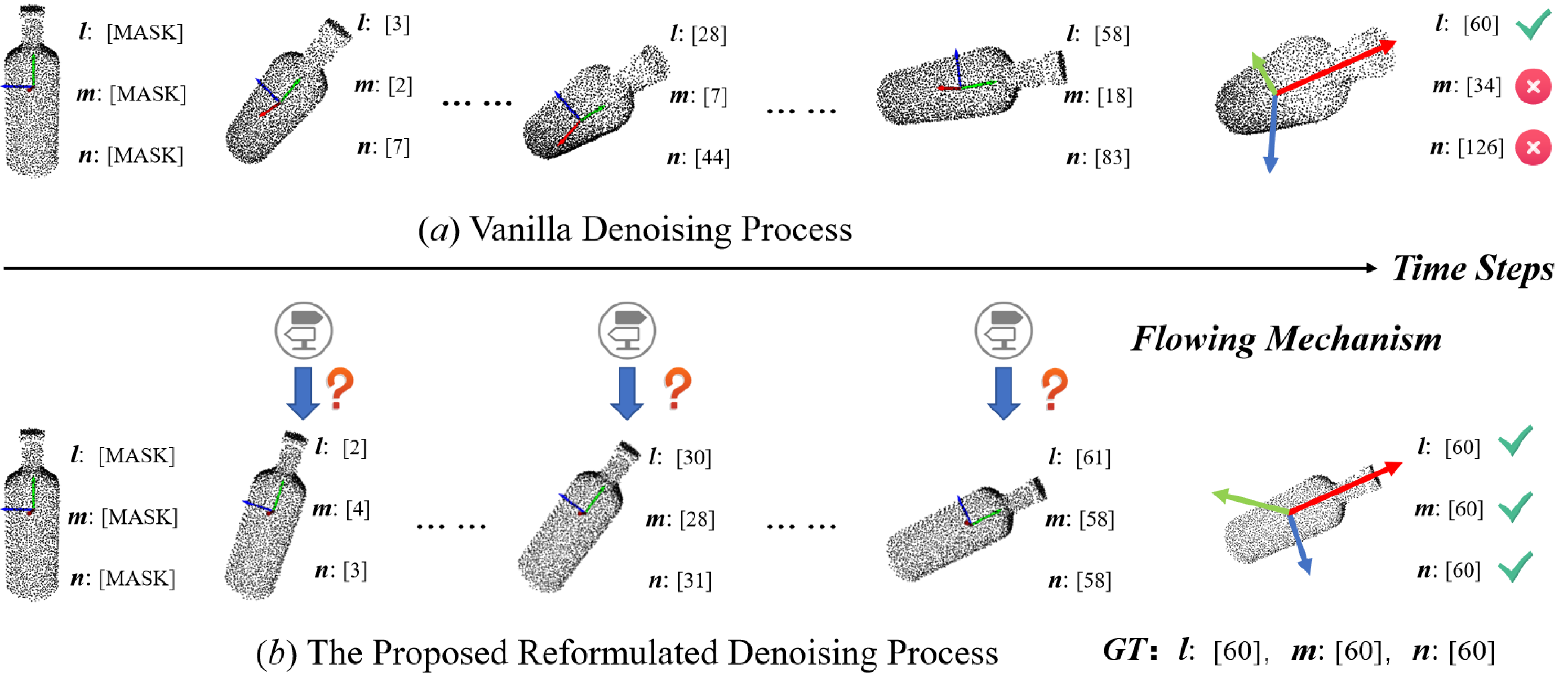}
    \caption{\textbf{Comparison of Different Denoising Processes.} We denote the rotation-related Euler angles as $l, m, n$, and model them using discretized bin indices for prediction. (a) illustrates the \textbf{vanilla denoising process} of conventional discrete diffusion models, where inconsistent convergence rates across tokens often introduce uncertainty and ambiguity in pose prediction—this can be viewed as an \textit{\textbf{aggressive}} denoising strategy. (b) presents the \textbf{reformulated denoising process} proposed in this work, which is centered around a customized \textit{Flowing Mechanism}. This mechanism introduces \textit{adaptive directional guidance} that determines appropriate update paths for each token. It is designed to enforce consistent convergence trajectories among semantically correlated token groups, thereby enabling a more stable and \textit{\textbf{gentle}} denoising process. Note that a rigid part (\textit{bottle}) is chosen to illustrate the process for simplicity.}
    \label{fig:teaser}
\end{figure*}

Compared to instance-level pose estimation, category-level pose estimation requires accurately predicting the 3D rotation and 3D translation of unseen objects, with the given partial observations. One approach~\cite{zhang2023genpose}, for example, utilizes the classic continuous diffusion model to perform pose estimation within a canonical space, representing objects of the same category. 
Despite significant progress in rigid object pose estimation, methods for articulated pose estimation have lagged behind, current methods are still facing several limitations:

\begin{itemize}
    \item Conventional pose regression methods~\cite{zhang2023genpose, zhang2025gapt, holmquist2023diffpose, wang2023posediffusion} rely on continuous space representations and often require exhaustive searches over large continuous domains for accurate estimation. Moreover, point clouds are discretely and non-uniformly sampled, leading to a mapping mismatch: discrete spatial inputs versus continuous pose outputs. This discrepancy poses a fundamental challenge, hindering precise modeling and limiting prediction performance.
    \item Prior methods studied on articulated object pose estimation (e.g., ~\cite{li2020category,liu2022akb,che2024op,zhang2024u}) tend to adopt a part-wise approach for pose modeling, where each part’s pose is estimated independently, disregarding the constraints imposed by the kinematic structure. Moreover, these methods exhibit limited robustness in addressing self-occlusion, especially when larger components obscure smaller movable parts from certain camera angles.
\end{itemize}

To tackle the first challenge, our core idea is to model articulation pose estimation as a discrete diffusion process. In this framework, the articulation pose is sampled by reversing a forward diffusion process that progressively transforms the real data into a series of noisy latent variables via a fixed Markov Chain. The reverse process starts from noise and iteratively denoises it, learning the posterior distribution in the process. Building on this discrete diffusion framework, we propose a novel yet equivalent re-formulation for the reverse process. To improve the balance between noisy and real distribution at various stages, we introduce the flexible flow decider (Fig.~\ref{fig:teaser}), which allows each token in the sequence to either be denoised or reset to a noisy state.

To address the second challenge, our solution is a hierarchical kinematic coupling  mechanism that defines part pairs. Conceptually, the rigid components of an articulated object can be divided into two categories: parent part, which can move in any direction (e.g., the main body of a cabinet) without kinematic constraints (typically, there is only one parent part), and child parts, which move along pre-defined joints (e.g., cabinet doors or drawers) according to physical kinematic laws (an articulated object may have multiple child parts). Leveraging this distinction, our approach represents each part's pose as a coupling state. This formulation is more amenable to learning by neural networks than direct 6D pose regression of each rigid part. Furthermore, the coupling state representation enhances pose estimation for occluded parts, as even limited visibility can provide sufficient information for accurate coupling state prediction.

Our contributions can be summarized as follows:

\begin{itemize}
\item DICArt is a novel framework designed to address category-level 6D pose estimation of articulated objects by formulating the articulation pose estimation as a discrete diffusion probabilistic process. 
\item To address challenges such as sub-optimal performance and articulation modeling, DICArt incorporates key modules, including the discrete diffusion probabilistic model, reformulated reverse process, and hierarchical kinematic coupling-guided pose estimation mechanism.
\item Extensive experiments across synthetic, semi-synthetic, and real-world datasets demonstrate that DICArt outperforms existing state-of-the-art methods on the articulation pose estimation (APE) task.
\end{itemize}

\section{Related work} \label{sec:Related_work}

$\bigstar$ \textbf{(1) Category-level 6D Pose Estimation.} 
Traditional instance-level pose estimation~\cite{doosti2020hope} aligns known 3D CAD models with observed objects. More recently, research has shifted toward category-level 6D pose estimation, which predicts an object’s 3D rotation and translation relative to a canonical category-specific representation. A seminal work, NOCS~\cite{wang2019normalized}, introduced normalized object coordinates and employed the Umeyama algorithm~\cite{umeyama1991least} for pose recovery. CASS~\cite{chen2020learning} later proposed a learned canonical shape space. To address intra-class shape variations, several methods have incorporated point-based shape priors~\cite{zou20226d}.
While most prior efforts focus on rigid objects, recent studies have begun addressing articulated objects~\cite{zhang2025r}, which consist of multiple rigid parts connected by joints. A-NCSH~\cite{li2020category} extended normalized coordinates to handle articulation, and Liu et al.~\cite{liu2022toward} further adapted this framework to real-world settings using part-pair reasoning to generalize to unseen instances.

\textit{Nevertheless, these approaches still face unresolved challenges, such as the necessity for explicit structural prior representations and issues arising from self-occlusions, making them less effective compared to our method.}

\noindent $\bigstar$ \textbf{(2) Diffusion Probabilistic Models.}

The diffusion generative model, originally proposed in~\cite{sohl2015deep}, has inspired extensive follow-up research~\cite{yang2023diffusion}. It comprises a forward Markov process that progressively adds noise to data and a reverse process that denoises it to recover the original distribution. Compared to autoregressive models, diffusion models offer advantages such as stable training and parallel refinement. Neural network-based reverse process learning has enabled impressive performance in continuous-state applications, including image generation~\cite{ho2022cascaded, rombach2022high}, editing~\cite{kawar2023imagic, couairon2022diffedit}, text synthesis~\cite{gu2022vector, ruiz2023dreambooth}, and audio processing~\cite{liu2023audioldm, alexanderson2023listen}.
In discrete domains, diffusion models naturally handle discrete variables. The original formulation in~\cite{sohl2015deep} considered binary variables, while~\cite{hoogeboom2021argmax} extended the approach to categorical data using uniform transition matrices. Further advancing this line,~\cite{austin2021D3PM} introduced Discrete Denoising Diffusion Probabilistic Models (D3PMs) with structured categorical corruption, improving modeling fidelity in discrete state spaces.

\textit{Existing studies in pose estimation task~\cite{wang2019normalized,chi2021garmentnets, wang2024dipose} have demonstrated the strong potential of discrete diffusion models in pose estimation tasks, primarily by constraining the generative process within physically plausible pose spaces to enhance prediction accuracy and stability. Inspired by this insight, we propose a novel design of the state transition matrix coupled with a tailored noise scheduling strategy, aiming to achieve a smoother and more progressive perturbation process that better aligns with the structural continuity and geometric plausibility required in pose estimation.}

\section{Notation and Problem Statement}
To achieve a robust algorithm for Category-level \textbf{A}rticulation \textbf{P}ose \textbf{E}stimation (\textbf{APE}) task, our key idea is to advance category-level articulation pose estimation in discrete state-spaces.
Here, we formulate a new paradigm for the APE task named DICArt. 
Concretely, given the partial observation $ \mathcal{P} = \{\delta _k\}_{k=1}^{K}$ as \textbf{\textit{input}} ($\{\delta_k\}$ is the $k$-th rigid part), DICArt conducts the predictions for \textit{\textbf{\rmnum{1}}}) per-part 3D rotation $R^{(k)} \in SO(3)$. \textit{\textbf{\rmnum{2}}}) per-part 3D translation $\mathbf{t}^{(k)}$.
In total, these together constitute the final pose estimation result:  $\mathcal{T} = \{R^{(k)}, \mathbf{t}^{(k)}\}_{k=1}^{K} \in SE(3)$ (\textbf{\textit{output}}).

In this paper, DICArt formulates the APE task as conditional generation and learns $p_\theta(\mathcal{T}|\mathcal{P})$, where $\mathcal{P} $ and $\mathcal{T} \in SE(3)$ refer to the input partial observation and target articulated object pose separately. For simplification, $\mathbf{x}$ is represented as the latent variable (i.e., articulation pose), which is composed of a set of pose elements $\boldsymbol{e}_i$. 
Each pose element $\boldsymbol{e}_i$ has three rotation-related tokens and three translation-related tokens. 
Each element will be processed as a sequence with 6 discrete tokens via bins.
i.e., $\boldsymbol{e}_i = \left\{l_i,m_i, n_i, x_i, y_i, z_i\right\}$, which are uniformly discretized into integers between $[1, \mathcal{K}]$. Then, we represent the pose as a concatenation of element sequences:

\begin{multline}
    \mathbf{x}  = \{ l_1,m_1, n_1, x_1, y_1, z_1 \Vert \dots \\ \Vert l_\kappa,m_\kappa, n_\kappa, x_\kappa, y_\kappa, z_\kappa \Vert \dots \|l_\mathcal{K},m_\mathcal{K}, n_\mathcal{K}, x_\mathcal{K}, y_\mathcal{K},z_\mathcal{K} \}
\end{multline}

Obviously, the pose sequence is \emph{heterogeneous}. Formally, the pipeline of our DICArt can be illustrated as follows: given the discrete  pose GT $\mathbf{x}_{0} \sim q\left(\mathbf{x}_{\mathbf{0}}\right)$, the \textit{forward} process corrupts it into a sequence of increasingly noisy latent variables $\mathbf{x}_{1: T}=\mathbf{x}_{\mathbf{1}}, \mathbf{x}_{2}, \ldots, \mathbf{x}_{T}$:

\begin{align}\label{eq:forward}
    & q(\mathbf{x}_{1:T}|\mathbf{x}_0)  = \prod_{t=1}^T q(\mathbf{x}_t|\mathbf{x}_{t-1}) \\\label{eq:transition}
    & q(\mathbf{x}_{t}|\mathbf{x}_{t-1}) = \mathbf{x}_t \mathbf{Q}_t \mathbf{x}_{t-1} \\
     & q(\mathbf{x}_t | \mathbf{x}_0) = \alpha_t\mathbf{x}_{t-1} + (1 - \alpha_t)q_\text{noise}
\end{align}

\noindent Where $\mathbf{x}_t$ denotes the one-hot version of a single discrete token in the pose sequence $\mathbf{x}_t$. $\alpha_t \coloneqq \prod_{i=1}^t \beta_i$ is specified to decrease from 1 to 0 w.r.t. $t$. $q_{\text{noise}}$ characterizes different diffusion processes, $\mathbf{Q}_t$ is the transition matrix, where $[\mathbf{Q}_t]_{ij}=q(\mathbf{x}_t=j|\mathbf{x}_{t-1}=i)$ represents the probabilities that $\mathbf{x}_{t-1}$ transitions to $\mathbf{x}_t$. Due to the property of Markov chain, the cumulative probability of $\mathbf{x}_t$ at arbitrary timestep from $\mathbf{x}_0$ can be derived as $q(\mathbf{x}_{t}|\mathbf{x}_{0})=\mathbf{x}_t \overline{\mathbf{Q}}_t \mathbf{x}_{0}$, where $\overline{\textbf{Q}}_t=\textbf{Q}_1\textbf{Q}_2\dots\textbf{Q}_t$. 
To conduct the pose prediction, the \emph{reverse} process starts with a random noise $\mathbf{x}_T$ and gradually recovers it relying on the learned posterior distribution $p_\theta(\mathbf{x}_{t-1}|\mathbf{x}_t)$, i.e.,  $p_\theta(\mathbf{x}_{0:T}) = p(\mathbf{x}_T)\prod_{t=1}^T p_\theta (\mathbf{x}_{t-1}|\mathbf{x}_t)$.

It is noted that a reformulated denoising process is introduced in this work, aiming at a gentle pose prediction. Finally, the parent part 6D pose, and axis descriptors are utilized to recover the final 6D pose via the spatial kinematic reasoning. To avoid ambiguity and better clarify the variables, we use the symbol ` $*$ ' to indicate the GT variables, and ` $\mathbf{\hat{}}$ ' to indicate the predicted variables.

\section{Methodology} \label{sec:Methodology}

\begin{figure*}[t!]
    \centering
    \includegraphics[width=0.88\linewidth]{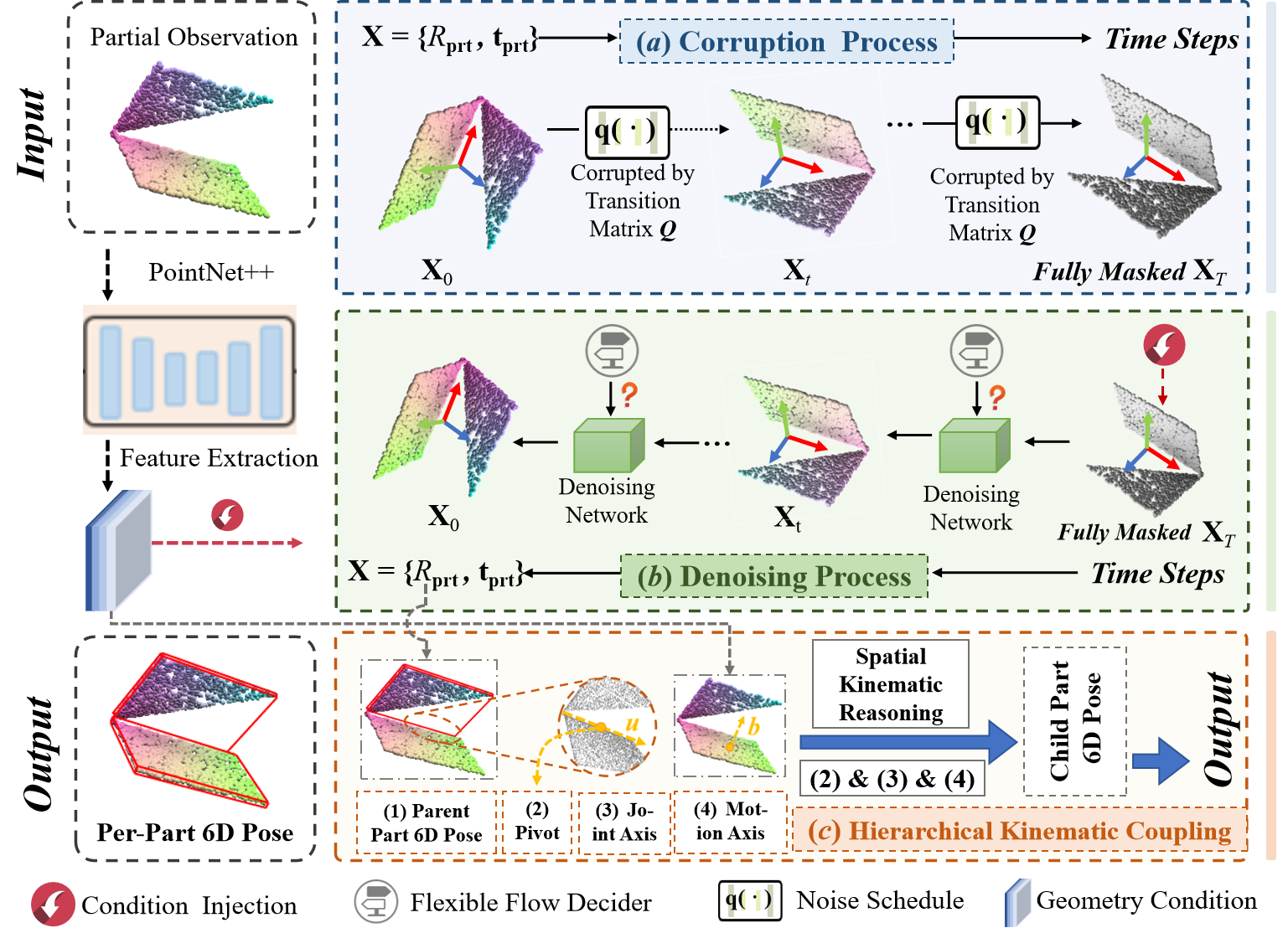}
    \caption{\textbf{The Pipeline of Our DICArt.} Please note that images of the object with varying saturation levels represent different degrees of noise in the pose annotations.}
    \label{fig:pipeline}
\end{figure*}

\subsection{Forward Corruption of Parent Part's Pose}\label{sec:Forward}
As illustrated in Fig.~\ref{fig:pipeline}, we first predict parent part pose $\mathcal{T}_{\text{prt}}=[R_{\text{prt}}~\vert~\mathbf{t}_{\text{prt}}]$ via discrete diffusion models under the \textit{hierarchical kinematic coupling} strategy. 
However, directly applying a discrete diffusion model is not straightforward. Inspired by prior works~\cite{wang2019normalized, chi2021garmentnets}, we reformulate the translation regression task $\mathbf{t} \in \mathbb{R}^3$ as a classification problem to enhance stability and convergence.
Specifically, each axis in the 3D coordinate space is discretized into a set of bins, and the network is trained to predict the index of the corresponding bin. Similarly, the rotation matrix is converted into three independent Euler angles, which are discretized within their domain (i.e., $[0^\circ, 360^\circ)$ or an equivalent periodic space), thereby transforming the rotation prediction into a classification task as well. This formulation results in two semantically distinct token groups: $\{x, y, z\}$ are translation-related tokens, and $\{l, m, n\}$ are rotation-related tokens, which facilitate structured and interpretable 6D pose prediction.

The design of the transition matrix is inspired by the following important observations:

(1) \textbf{Token Consistency}. The pose sequence consists of two semantically distinct components—rotation and translation, each comprising three dimensions, resulting in a heterogeneous mixture of six token types. Directly applying conventional discrete diffusion models for state transitions may lead to semantic inconsistencies—for instance, rotation-related tokens may be erroneously transformed into translation-related tokens, thereby compromising the structural integrity and physical meaning of the pose representation. To address this issue, we introduce a block-wise diagonal constraint in the state transition matrix, which isolates different token types into separate subspaces and restricts transitions to occur only within tokens of the same semantic category. This design effectively ensures both the validity of sequence generation and the stability of pose prediction. 
Therefore, to ensure the validity of the predicted pose sequence, we restrict internal transitions to occur only within the same type of token. Consequently, $\mathbf{Q}_t^{pose}$ simplifies to a block-wise diagonal matrix, represented mathematically as:

\begin{equation} \label{eq:transition_all}
    \mathbf{Q}_t^{pose}={
\left[ \begin{array}{cc}
\mathbf{Q}_t^{rot} &\\
& \mathbf{Q}_t^{tsl}
\end{array} 
\right ]},
\end{equation}

\noindent Where $\mathbf{Q}_t^{rot}$ and $\mathbf{Q}_t^{tsl}$ represent transition probabilities of rotation and translation tokens, respectively.

(2) \textbf{Smooth Classification}. While discretization—such as binning—enables pose estimation to be cast as a classification problem, it differs significantly from traditional categorical classification.
To better capture this structural continuity, we draw inspiration from paper~\cite{austin2021D3PM}. Specifically, each token in the pose sequence evolves under a transition matrix $Q_t$ that models its probabilistic trajectory during denoising. To improve responsiveness to uncertain or noisy tokens, we extend the discrete state space from $\mathcal{K}$ to $\mathcal{K}+1$ by introducing a special [\texttt{MASK}] token. This augmentation enables $Q_t \in \mathbb{R}^{(\mathcal{K}+1) \times (\mathcal{K}+1)}$ to jointly encode $\mathcal{K}$ quantized pose classes and a dynamic masking state, thereby facilitating more flexible and effective correction of pose tokens during the reverse process.

\subsection{Reformulated Denoising Process}\label{sec:Reformulation}

As previously mentioned, we convert the rotation matrix into three semantically correlated yet independent Euler angles. Due to the strong interdependence among these angles, traditional discrete diffusion models~\cite{yang2023diffsound,chen2022analog} often struggle to denoise them at a synchronized pace during the reverse process: some tokens may prematurely converge to their true values before the final diffusion step (i.e., $t < T$), while others remain in a high-noise state. This asynchronous convergence disrupts the semantic consistency among tokens and ultimately degrades the accuracy of pose estimation. This limitation underscores the fundamental challenge faced by conventional discrete diffusion models when dealing with tightly coupled token representations. To address this issue, we revisit the reverse process and introduce a novel \textit{Reformulated Reverse Process}, which is specifically designed to maintain semantic coherence and synchronization during denoising, thereby enhancing the overall prediction accuracy and stability.

Our core idea is a \textit{flexible flow decider}, to apply it in the conditional reverse transition of discrete diffusion models, we draw inspiration from ~\cite{austin2021D3PM, hoogeboom2021argmax,zheng2023reparameterized}, where the variational approach demonstrates that the reverse process shuttles discrete tokens between noisy states and the ground truth state \( \mathbf{x}_0 \), conditioned on the equality of \( \mathbf{x}_t \) and \( \mathbf{x}_0 \). When \( \mathbf{x}_t = \mathbf{x}_0 \), the token is likely in a noise-free state. In this scenario, the model maintains the noise-free state by setting \( \mathbf{x}_{t-1} = \mathbf{x}_t \) or reintroduces noise. When \( \mathbf{x}_t \neq \mathbf{x}_0 \), the model treats the token as noisy and either denoises \( \mathbf{x}_t \) towards \( \mathbf{x}_0 \) or keeps it in a noisy state.
Concretely, given the forward transition matrix \( \left[\mathbf{Q}_t^{pose}\right] \), the transition formulation can be expressed compactly, i.e., the conditional reverse transition \( q(\mathbf{x}_{t-1} | \mathbf{x}_t, \mathbf{x}_0) \) is represented as:

\begin{small}
\begin{equation}
    q(\mathbf{x}_{t-1} | \mathbf{x}_{t}, \mathbf{x}_0) = \begin{cases}
        \lambda^{(1)}_{t-1}\mathbf{x}_t + \left(1-\lambda^{(1)}_{t-1}\right) \mathbf{x}_{T}, &\textup{ if } \mathbf{x}_t = \mathbf{x}_0 \\
        \lambda^{(2)}_{t-1}\mathbf{x}_0 + \left(1 - \lambda^{(2)}_{t-1}\right)q_\textup{noise}(\mathbf{x}_t), &\textup{ if } \mathbf{x}_t \neq \mathbf{x}_0.
    \end{cases}\label{eqn:q_xtm1_xt_x0_as_branching}
\end{equation}
\end{small}

The coefficient $\lambda^{(1)}_{t-1}, \lambda^{(2)}_{t-1}$ are the mapping results of $\alpha_{t-1}, \beta_{t-1}$, which governs the probability of transitioning directly from $\mathbf{x}_t$ to the ground truth state $\mathbf{x}_0$.
The terminal state $\mathbf{x}_T$ is associated with the [\texttt{MASK}] token, denoted as the noise distribution $q_{\text{noise}}$.

To integrate the sampling method through an augmented path and simplify the training process, we obtain the complete reformulation as follows:

\begin{equation}
\begin{aligned}
b_t &= \mathbb I \{\mathbf{x}_t = \mathbf{x}_0\}\\
v_{t-1}^{(1)} &\sim \text{GS}\left(\lambda^{(1)}_{t-1}, 1 - \lambda^{(1)}_{t-1} \right),\quad u_{t}^{(1)} \sim \mathbf{x}_T\\
v_{t-1}^{(2)} &\sim \text{GS}\left(\lambda^{(2)}_{t-1}, 1 - \lambda^{(2)}_{t-1}\right),\quad u_{t}^{(2)} \sim q_{\text{noise}}(\mathbf{x}_t)\\
\mathbf{x}_{t-1} &= b_t\left[v_{t-1}^{(1)}\mathbf{x}_t + \left(1-v_{t-1}^{(1)}\right) u_{t}^{(1)}\right] \\ &+ (1-b_t)\left[v_{t-1}^{(2)}\mathbf{x}_0 + \left(1-v_{t-1}^{(2)}\right) u_{t}^{(2)}\right] \label{eqn:reparam_backward}
\end{aligned}
\end{equation}

\noindent To simplify the notation, we denote $v_{t-1} \coloneqq \left[v_{t-1}^{(1)}, v_{t-1}^{(2)}\right]$,  $u_{t-1} \coloneqq \left[u_{t-1}^{(1)}, u_{t-1}^{(2)}\right]$ and $\lambda_{t-1} \coloneqq \left[\lambda^{(1)}_{t-1}, \lambda^{(2)}_{t-1}\right]$. 
In Eq~\ref{eqn:reparam_backward}, $\mathbb I$ is indicator function, $q_\textup{noise}(\mathbf{x}_t) = \beta_t \mathbf{x}_t + (1 - \beta_t) q_\textup{noise}$,  taking the value 1 when $\mathbf{x}_t = \mathbf{x}_0$ and 0 otherwise. The variable $v_{t-1}$, drawn from a Gumbel-Softmax function~\cite{jang2016categorical} (GS ($\cdot$)) with time parameter $\lambda_{t-1}$, governs the discrete \textit{flowing mechanism}, which determines the transition path at each step of the diffusion process. $u_{t}$ denotes  a noise distribution that sampled by $\mathbf{x}_T$ or  $\mathbf{x}_t$.

This framework extends the standard discrete diffusion model by introducing a series of binary flow indicators $\{v_{t-1}\}_{t=1}^T$, allowing the model to dynamically choose between preserving, denoising, or reintroducing noise at each timestep. The conditional reverse transition is computed by marginalizing over $v_{t-1}$, yielding:
\begin{equation}
    q(\mathbf{x}_{t-1} \mid \mathbf{x}_t, \mathbf{x}_0) = \mathbb{E}_{v_{t-1} \sim \text{GS}(\lambda_{t-1})} \left[q(\mathbf{x}_{t-1} \mid v_{t-1}, \mathbf{x}_t, \mathbf{x}_0)\right].
\end{equation}
By explicitly modeling the uncertainty in token behavior through $v_{t-1}$, this joint formulation enhances both the flexibility and representational capacity of the diffusion process.

\subsection{Hierarchical Kinematic Coupling} \label{sec:Joint-Oriented}

To further reduce the pose prediction space (the core motivation of this work) and to address the prevalent issue of physical inconsistency in existing methods, we introduce kinematic reasoning and a hierarchical coupling mechanism into the part pose estimation process, forming the \textit{hierarchical kinematic coupling} manner for articulated objects. The core idea of this paradigm lies in the fact that an articulated object is composed of multiple rigid components whose motions are coupled through joint axes within a hierarchical structure. Each sub-part’s motion trajectory is strictly constrained by its corresponding joint axis rather than varying arbitrarily.

Based on this property, we categorize the rigid components of an articulated object into two types: 1) \textit{Parent Part}, which serving as the motion reference of the entire articulated structure, it can move freely in 3D space (typically, an articulated object contains only one parent part, such as the main body of a cabinet); 2) \textit{Child Part}, whose motion strictly depends on the parent part and the associated joint, being confined to directions or trajectories defined by the joint and constrained by the underlying kinematic structure (e.g., cabinet doors or drawers).

According to the kinematic principles of articulated objects, joint axes can be categorized into two types, each defined by specific descriptors: 1) Revolute Axis: enables the child part to rotate around a fixed axis, as exemplified by a laptop’s hinge connecting the base (parent) and the screen (child). Its axis descriptor is parameterized as $\boldsymbol{\phi}_{r} = (\boldsymbol{u}_{r}, \boldsymbol{q}_{r})$, where $\boldsymbol{u}_{r} \in \mathbb{R}^3$ denotes the unit direction vector of the joint axis in the canonical coordinate system, and $\boldsymbol{q}_{r} \in \mathbb{R}^3$ represents the pivot point, i.e., the rotation center. 2) Prismatic Axis: allows linear motion of the child part along a specific direction, such as a drawer sliding relative to its cabinet. It is parameterized as $\boldsymbol{\phi}_{p} = (\boldsymbol{u}_{p})$, where $\boldsymbol{u}_{p}\in \mathbb{R}^3$ denotes the unit direction vector of the joint axis.
This unified parameterization establishes a consistent mathematical foundation for hierarchical kinematic coupling–based pose estimation.

To precisely model the joint constraint information of articulated objects, two independent MLPs are designed to predict the axis descriptor and the motion axis $\boldsymbol{b}^{(k)}$, respectively. Specifically: 1) Axis descriptor prediction: this module aims to determine the joint axis direction $\boldsymbol{u}$. The alignment axis direction of each child part, denoted as $\boldsymbol{a}^{(k)}$, is defined to be parallel to $\boldsymbol{u}$. The predicted $\boldsymbol{u}$ serves as the initialization basis and is further mapped to $\boldsymbol{a}^{(k)}$, ensuring the spatial geometric consistency and interpretability of the joint axis orientation. 2) Motion axis prediction: to enhance the stability and physical plausibility of joint axis estimation, an orthogonality constraint  is imposed, enforcing the motion axis $\boldsymbol{b}^{(k)}$ to be strictly orthogonal~\cite{di2022gpv} to the joint axis $\boldsymbol{a}^{(k)}$  in 3D space. This motion axis defines the relative motion direction of the child part with respect to its parent part, ensuring that the predicted motion trajectory complies with the kinematic rules and physical constraints of articulated motion.

\begin{table*}[!t]
    \centering
    \caption{\textbf{Comparisons on the ArtImage Dataset.} We validate our DICArt on five categories (Laptop, Eyeglasses, Dishwasher, Scissors, and Drawer) that contain 2, 3, 2, 2, 4 parts, respectively. }
    \resizebox{\linewidth}{!}{
    \begin{tabular}{c|l|cc|c|cc}
    \toprule
    \multirow{2}{*}{Category} & \multirow{2}{*}{Method} & \multicolumn{2}{c|}{Per-part 6D Pose}  &\multicolumn{2}{c}{Axis Descriptor} \\
    \cline{3-6}
    & & Rotation Error ($^{\circ}$) $\downarrow$ & Translation Error (m) $\downarrow$  & Angle Error ($^{\circ}$) $\downarrow$ & Distance Error (m) $\downarrow$\\
    \hline
    \multirow{5}{*}{Laptop} & A-NCSH \cite{li2020category} & 5.3, 5.4 & 0.054, 0.043  & 1.7&0.09  \\
    & GenPose~\cite{zhang2023genpose} & 5.3, 6.1 & 0.068, 0.060 & 3.8&\textbf{0.03}  \\
    & OP-Align~\cite{che2024op} & 6.3, 5.2 & 0.072, 0.058 & 4.5 & 0.11 \\
    & ShapePose~\cite{zhou2025canonical} & 5.0, 4.6 & 0.052, 0.064 &3.3&0.06 \\
    & \textbf{DICArt (Ours)} & \textbf{3.2}, \textbf{3.9} &\textbf{0.045}, \textbf{0.040}  & \textbf{0.8}&0.04 \\
    \hline
    
    \multirow{5}{*}{Eyeglasses} & A-NCSH \cite{li2020category} & 3.7, 22.3, 23.2 & 0.049, 0.313, 0.324 &3.1, 3.1&0.07, 0.06 \\
    & GenPose~\cite{zhang2023genpose} & 5.0, 7.4, 7.6 & 0.063, 0.113, 0.301 &4.1, 4.3 & 0.04, 0.05 \\
    & OP-Align~\cite{che2024op} & 9.3, 8.2, 15.6  & 0.048, 0.253, 0.281 & 2.5, 3.2 & 0.06, 0.08 \\
    & ShapePose~\cite{zhou2025canonical} & 4.2, 6.0, 6.0 & 0.049, 0.106, 0.108 &3.8, 3.9&0.05, 0.08 \\
    & \textbf{DICArt (Ours)} & \textbf{3.5}, \textbf{5.1}, \textbf{5.3} & \textbf{0.041}, \textbf{0.091}, \textbf{0.083} &\textbf{1.5}, \textbf{1.6}&\textbf{0.02}, \textbf{0.02} \\
    \hline
    
    \multirow{5}{*}{Dishwasher} & A-NCSH \cite{li2020category} & 4.0, 4.8 & 0.059, 0.123 &6.1&0.11 \\
    & GenPose~\cite{zhang2023genpose} & 6.1, 6.3 & 0.115, 0.164 &4.8&0.09 \\
     & OP-Align~\cite{che2024op} & 4.2, 5.8 & 0.132, 0.182 & 5.3 & 0.10 \\
    & ShapePose~\cite{zhou2025canonical} & 3.9, 4.3 & 0.055, 0.079  &2.2&0.04 \\
    & \textbf{DICArt (Ours)} & \textbf{2.9}, \textbf{3.7} & \textbf{0.048}, \textbf{0.055} &\textbf{1.7}&\textbf{0.03} \\
    \hline
    
    \multirow{5}{*}{Scissors} & A-NCSH \cite{li2020category} & 2,0, 2.9 & 0.035, 0.025 &0.8&0.04 \\
    & GenPose~\cite{zhang2023genpose} & 4.1, 3.5 & 0.050, 0.041 &2.8&0.06 \\
     & OP-Align~\cite{che2024op} & 3.3, 3.7 & 0.043, 0.058 & 2.5 & 0.09 \\
    & ShapePose~\cite{zhou2025canonical} & 2.3, 2.9 & 0.033, 0.045  &1.9&0.08 \\
    & \textbf{DICArt (Ours)} & \textbf{1.7}, \textbf{2.2} & \textbf{0.024}, \textbf{0.021} &\textbf{0.5}&\textbf{0.02}\\
    \hline
    
    \multirow{5}{*}{Drawer} & A-NCSH~\cite{li2020category} & 2.8, 3.5, 3.9, 2.9 & 0.045, 0.155, 0.157, 0.075 &2.6, 2.7, 5.2&-, -, -\\
    & GenPose~\cite{zhang2023genpose} & 4.3, 4.3, 4.3, 4.3 & 0.125, 0.154, 0.166, 0.126&3.3, 3.3, 3.3&-, -, - \\
    & OP-Align~\cite{che2024op} & 5.3, 4.5, 4.6, 4.9  & 0.101, 0.137, 0.178, 0.135 & 2.9, 3.1, 3.0  &-, -, -\\
    & ShapePose~\cite{zhou2025canonical} & 3.6, 3.6, 3.6, 3.6 & 0.068, 0.126, 0.128, 0.115 &2.0, 2.3, 2.1&-, -, - \\
    & \textbf{DICArt (Ours)} & \textbf{1.7}, \textbf{1.7}, \textbf{1.7}, \textbf{1.7} & \textbf{0.041}, \textbf{0.085}, \textbf{0.091}, \textbf{0.071}  &\textbf{1.4}, \textbf{1.8}, \textbf{1.9} &-, -, - \\
    \bottomrule
    \end{tabular}}
    \label{tab:main_table}

\end{table*}

\begin{figure*}[t!]
    \centering
    \includegraphics[width=0.98\linewidth]{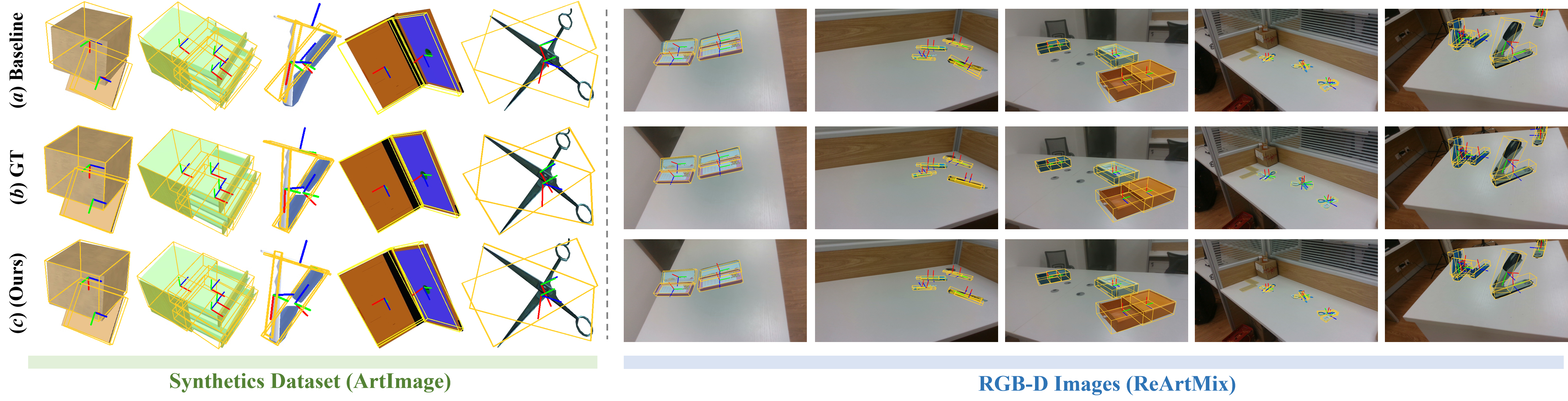}
    \caption{\textbf{Qualitative Results on the Synthetic
Dataset (left) and RGB-D Images Dataset (right).}}
       \label{fig:main_results}
\end{figure*} 

\section{Experiments} \label{sec:Experiments}
\subsection{Experimental Setup}
\textbf{Dataset, Baselines, Metrics.} The following datasets are used: synthetic dataset, ArtImage~\cite{xue2021omad}, semi-synthetic dataset ReArtMix~\cite{liu2022toward}, and real-world dataset RobotArm \cite{liu2022toward}.  To ensure a fair and comprehensive comparison, the following baselines are selected: A-NCSH~\cite{li2020category}, GenPose~\cite{zhang2023genpose}, OP-Align~\cite{che2024op}, ShapePose~\cite{zhou2025canonical}. To validate the performance of the DICArt, we adopt degree error~($^\circ$) for 3D rotation, distance error~(m) for 3D translation. 

\noindent \textbf{Experimental Details.} 
During the data preprocessing phase, point clouds are uniformly sampled to contain exactly 1,024 points, which serve as the inputs to the network. We utilize the Adam optimizer, the learning rate is 0.001, the batch size is 96, and the total number of training epochs is 200. The total diffusion steps $T$ is 100, and the bin size is 360. All experiments were conducted on an NVIDIA GeForce RTX 3090 GPU equipped with 24GB of memory.

\subsection{Comparison with the SOTA Methods}

\textbf{Quantitative Results.} 
We present the results of DICArt on ArtImage in Tab.~\ref{tab:main_table}. Compared to classical methods, we achieve the best pose estimation results for the \textit{laptop} category, with rotation errors of $\mathbf{3.2}^{\circ}$ and $\mathbf{3.9}^{\circ}$. This superior performance can be attributed to \textit{reformulated reverse process}, which excels in handling objects with similar sizes and shapes at the part level. For the \textit{Eyeglasses} category, our method demonstrates exceptional performance, achieving translation errors of \textbf{0.041}$m$, \textbf{0.091}$m$, and \textbf{0.083}$m$, which highlights the framework's precision in estimating axis descriptors. Concerning pose error of child part, we achieve a remarkable $\mathbf{3.3}^{\circ}$ and $\mathbf{0.05}m$ error in average for category \textit{dishwasher}, indicating that articulation modeling with a hierarchical kinematic coupling perspective is more effective than part-wise approaches.

\noindent  \textbf{Qualitative Results.}   
In Fig.~\ref{fig:main_results}, we show the visualizations of Ground Truth, A-NCSH and our method on ArtImage~(left). It can be observed that, compared to A-NCSH, our method demonstrates superior performance in predicting the pose of each part of the articulated object in camera space. Specifically, our method achieves results that are very close to the GT for objects such as laptops, dishwashers, and drawers. This indicates that our approach not only provides more accurate predictions but also maintains high consistency across different types of articulated objects.

\subsection{Ablation Study}

\begin{table}[t!] 
\centering
 \caption{\textbf{Ablation Study Results}. Experiments are conducted on the category \textit{Drawer}.}
\resizebox{0.95\linewidth}{!}{
	\begin{tabular}{c|ccc|c|c}
            \toprule

\textbf{Index} & \multicolumn{3}{c|}{\textbf{Diffusion}}   & \textbf{Rotation Error ($^{\circ}$)}   & \textbf{Translation Error ($m$)}   \\
            \midrule
\Rmnum{1}  & \multicolumn{3}{c|}{Continuous}   &  3.1   & 0.143  \\
\Rmnum{2}  & \multicolumn{3}{c|}{Discrete}   &  1.7   & 0.072  \\

\midrule
\midrule

\textbf{Index} & \multicolumn{3}{c|}{\textbf{Occlusion Level (Visibility)}}   & \textbf{Rotation Error ($^{\circ}$)}   & \textbf{Translation Error ($m$)}   \\
            \midrule
\Rmnum{3}   & \multicolumn{3}{c|}{0\%-40\%}   & 1.8   & 0.089  \\
\Rmnum{4}  & \multicolumn{3}{c|}{40\%-80\%}   & 1.9   & 0.096  \\
\Rmnum{5}  & \multicolumn{3}{c|}{80\%-100\%}  & 1.9  & 0.107  \\

\midrule
\midrule

\textbf{Index} & \multicolumn{3}{c|}{\textbf{Reformulated Denoising}}   & \textbf{Rotation Error ($^{\circ}$)}   & \textbf{Translation Error ($m$)}   \\
            \midrule
\Rmnum{6}   & \multicolumn{3}{c|}{-}   & 4.0   & 0.128  \\
\Rmnum{7}  & \multicolumn{3}{c|}{\checkmark}  &   1.7 & 0.072  \\

\bottomrule
	\end{tabular} 
 }
\label{tab:ablation}
\end{table}

\textbf{Discrete Diffusion \textit{v.s} Continuous Diffusion.} In this paper, we present a novel and effective method for APE task using a discrete diffusion model, as opposed to the continuous diffusion model. We argue that pose estimation in discrete state spaces is more effective in the pre-configured state space rather than in a continuous one. 
To validate the effectiveness of our approach, DICArt, we replace the discrete process in DICArt with a continuous one~\cite{zhang2023genpose} as the baseline. The comparison results, shown in Tab.~\ref{tab:ablation} (\Rmnum{1} - \Rmnum{2}), demonstrate the superiority of our method, which significantly outperforms the baseline in terms of performance.

\noindent \textbf{Self-occlusion Analysis.} To further examine the robustness of our DICArt under self-occlusion conditions, we categorized the test samples of the \textit{Drawer} category into three subsets according to their occlusion levels, as self-occlusion issues are particularly prevalent in drawers. The occlusion level is quantified by the visible ratio of points relative to the total number of points, a metric akin to that used in A-NCSH~\cite{li2020category}. Specifically, this experiment involves three subsets characterized by different visibility ratios: (0\%, 40\%), (40\%, 80\%), and (80\%, 100\%). We utilized both mean rotation and translation errors as uniform metrics to assess the pose estimation performance. As shown in Tab.~\ref{tab:ablation}~(\Rmnum{3}-\Rmnum{5}), the results indicate that, despite increasing occlusion levels, the rotation error remains largely stable, a testament to the effectiveness of our hierarchical kinematic coupling approach. Notably, our method maintains a low translation error even at occlusion levels up to 80\%. Even in extreme scenarios where occlusion ranges from 80\% to 100\%, our method continues to deliver satisfactory results, with a translation error of merely 0.107$m$.

\noindent  \textbf{Reformulated Denoising.} As described in Sec.~\ref{sec:Reformulation}, traditional diffusion models often struggle to efficiently recover all tokens from pure noise to the target data distribution in a consistent manner during the denoising process. To address this limitation, we introduce the \textit{Flowing Mechanism}, which centers on the proposed \textit{Flexible Flow Decider}. This component adaptively guides the update direction and magnitude of each token at every denoising step. Ablation studies presented in Tab.~\ref{tab:ablation}~(\Rmnum{6}-\Rmnum{7}) demonstrate that the proposed method significantly enhances model performance, owing to its \textit{gentle yet adaptive} denoising strategy, which effectively mitigates the imbalance in convergence commonly observed in traditional diffusion processes.

\subsection{Generalization Capacity}

\begin{table}[t!] 
\centering
\caption{\textbf{Pose Estimation Results on ReArtMix Dataset}.}
\resizebox{0.95\linewidth}{!}{
\begin{tabular}{c c cc}
\toprule
\multirow{2}{*}{Category} & \multirow{2}{*}{Method} & \multicolumn{2}{c}{Per-part Pose} \\
\cline{3-4}
& & Rotation Error ($^\circ$) & Translation Error (m)  \\
\hline
\multirow{2}{*}{Box} & ReArtNet~\cite{liu2022toward} & 3.3, 3.4 & 0.026, 0.031  \\
& DICArt~(Ours)  & \textbf{1.5}, \textbf{1.6} & \textbf{0.009}, \textbf{0.010}  \\
\hline
\multirow{2}{*}{Stapler} & ReArtNet~\cite{liu2022toward} & 5.1, 6.5 & 0.035, 0.031  \\
& DICArt~(Ours)  & \textbf{2.5}, \textbf{3.2} & \textbf{0.008}, \textbf{0.012} \\
\hline
\multirow{2}{*}{Cutter} & ReArtNet~\cite{liu2022toward} & 3.3, 3.0 & 0.019, 0.014 \\
& DICArt~(Ours)  & \textbf{2.3}, \textbf{2.4} & \textbf{0.007}, \textbf{0.009} \\
\hline
\multirow{2}{*}{Scissors} & ReArtNet~\cite{liu2022toward} & 5.4, 5.1 & 0.009, 0.015  \\
& DICArt~(Ours)  & \textbf{3.5}, \textbf{4.2} & \textbf{0.008}, \textbf{0.011} \\
\hline
\multirow{2}{*}{Drawer} & ReArtNet~\cite{liu2022toward} & 3.4, 3.9 & 0.024, 0.021 \\
& DICArt~(Ours)  & \textbf{1.5}, \textbf{1.6} & \textbf{0.007}, \textbf{0.009} \\
\bottomrule
\end{tabular}}
\label{tab:reart_exp}
\end{table}

\textbf{Experiments on Semi-Synthetic Scenarios.} Tab.~\ref{tab:reart_exp} shows quantitative results on the ReArtMix dataset~\cite{liu2022toward}. Compared to the baseline, our method demonstrates excellent performance, achieving superior rotation estimation across all categories. Specifically, our method exhibits highly accurate translation estimation, with translation errors of only \textbf{0.007}$m$ and \textbf{0.009}$m$ for the \textit{Drawer} category. This level of precision underscores the robustness and reliability of our approach in handling various types of articulated objects. Qualitative results can be seen in Fig.~\ref{fig:main_results}~(right).

\begin{table}[t!] 
\scriptsize
\caption{\textbf{Pose Estimation Results on 7-part RobotArm Dataset.}}
\small
\centering
\resizebox{\linewidth}{!}{
\begin{tabular}{c|ccccccc}
\toprule
\multicolumn{8}{c}{Per-part Rotation Error ($^{\circ}$)} \\
\hline
Part ID & 1 & 2 & 3 & 4 & 5 & 6 & 7 \\
\hline
A-NCSH \cite{li2020category} & 7.8 & 7.9 & 10.3 & 10.5 & 11.2 & 16.4 & 23.5 \\
DICArt~(Ours)  & \textbf{1.6} & \textbf{4.8} & \textbf{7.8} & \textbf{7.9} & \textbf{8.1} & \textbf{12.3} & \textbf{15.1} \\
\hline
\multicolumn{8}{c}{Per-part Translation Error (m)} \\
\hline
Part ID & 1 & 2 & 3 & 4 & 5 & 6 & 7 \\
\hline
A-NCSH \cite{li2020category} & 0.012 & 0.044 & 0.067 & 0.066 & 0.179 & 0.236 & 0.403 \\
DICArt~(Ours)  & \textbf{0.011} & \textbf{0.031} & \textbf{0.051} & \textbf{0.053} & \textbf{0.082} & \textbf{0.156} & \textbf{0.353} \\
\bottomrule
\end{tabular}}
\label{tab:robotarm_exp}
\end{table}

\begin{figure}[h!]
\centering
    \includegraphics[width=\linewidth]{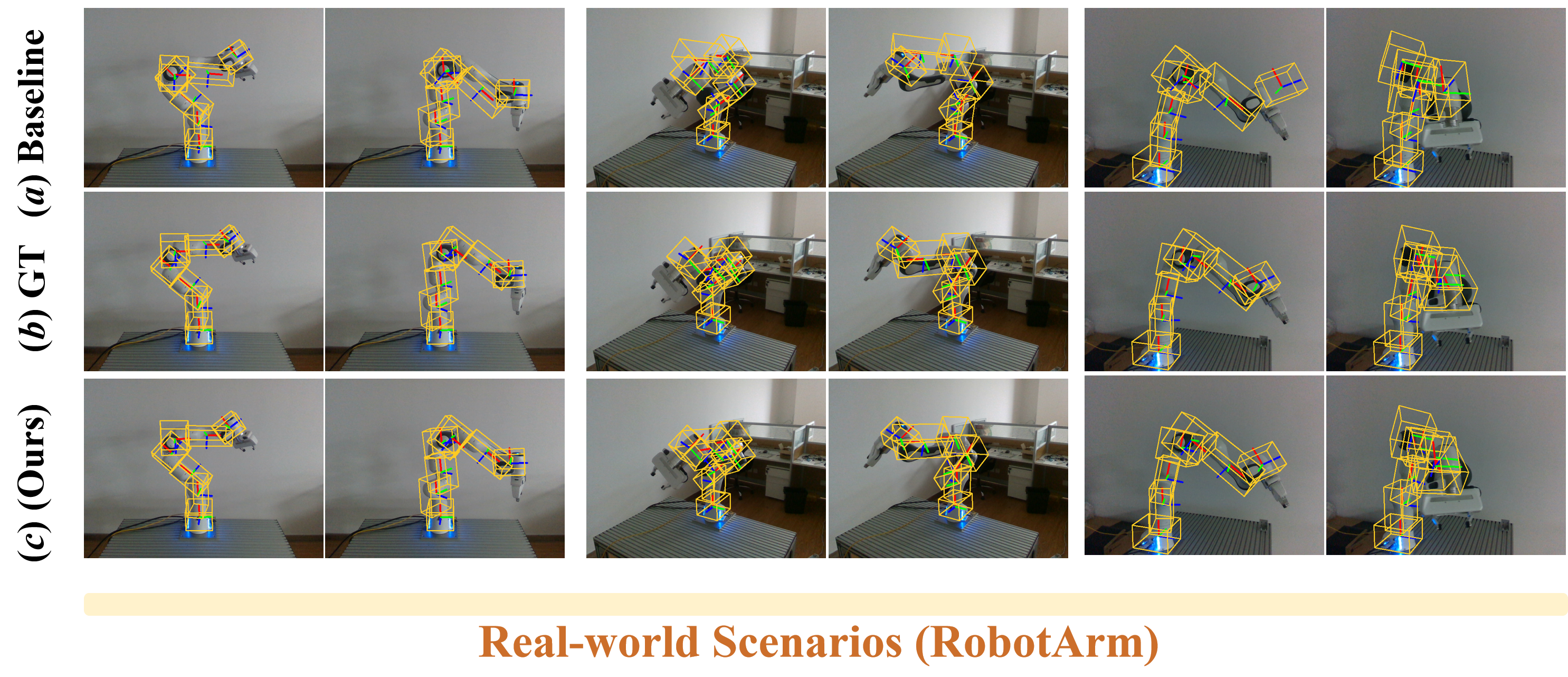}
    \caption{\textbf{Qualitative Results on 7-part RobotArm Dataset.}}
\label{fig:RobotArm}
\end{figure}

\noindent \textbf{Experiments on Real-world Scenarios.} We also evaluate DICArt using the 7-part RobotArm dataset in real-world scenarios. As shown in Tab.~\ref{tab:robotarm_exp}, our method performs well in estimating per-part poses, achieving an average rotation error of \textbf{8.2$^\circ$} and a translation error of \textbf{0.105$m$} across parts 1 to 7. These results demonstrate the robustness and accuracy of our approach in handling complex, multi-part articulations in real-world settings. Qualitative results are shown in Fig.~\ref{fig:RobotArm}, further confirming the effectiveness of our method.

\section{Conclusion}

In this paper, we propose DICArt, a novel category-level articulated object pose estimation framework via improved discrete diffusion models. The central idea is to introduce discretization, transforming the pose estimation task into a more manageable classification problem. To restrictively confine the search space towards physically viable configurations and a gentle denoising process, a reformulated denoising mechanism is introduced.
To cope with the self-occlusion problems, we propose the hierarchical kinematic coupling mechanism, which takes the kinematic constraints into consideration. 
Experimental results demonstrate that our method not only achieves SOTA performance on the synthetic dataset (ArtImage) but also exhibits strong generalization capabilities on semi-synthetic (ReArtMix) and real-world articulated object datasets (RobotArm).

\section*{Acknowledgment}
This work was supported in part by National Natural Science Foundation of China under Grant 62302143 and 52275514, also with Anhui Provincial Natural Science Foundation under Grant 2308085QF207.

\clearpage
{
    \small
    \bibliographystyle{ieeenat_fullname}
    \bibliography{main}

@article{yu2023gamma,
  title={GAMMA: Generalizable Articulation Modeling and Manipulation for Articulated Objects},
  author={Yu, Qiaojun and Wang, Junbo and Liu, Wenhai and Hao, Ce and Liu, Liu and Shao, Lin and Wang, Weiming and Lu, Cewu},
  journal={arXiv preprint arXiv:2309.16264},
  year={2023}
}

@article{jang2016categorical,
  title={Categorical reparameterization with gumbel-softmax},
  author={Jang, Eric and Gu, Shixiang and Poole, Ben},
  journal={arXiv preprint arXiv:1611.01144},
  year={2016}
}

@inproceedings{zhao2024wavelet,
  title={Wavelet-based fourier information interaction with frequency diffusion adjustment for underwater image restoration},
  author={Zhao, Chen and Cai, Weiling and Dong, Chenyu and Hu, Chengwei},
  booktitle={Proceedings of the IEEE/CVF Conference on Computer Vision and Pattern Recognition},
  pages={8281--8291},
  year={2024}
}

@inproceedings{zhao2025zero,
  title={From Zero to Detail: Deconstructing Ultra-High-Definition Image Restoration from Progressive Spectral Perspective},
  author={Zhao, Chen and Chen, Zhizhou and Xu, Yunzhe and Gu, Enxuan and Li, Jian and Yi, Zili and Wang, Qian and Yang, Jian and Tai, Ying},
  booktitle={Proceedings of the Computer Vision and Pattern Recognition Conference},
  pages={17935--17946},
  year={2025}
}

@article{zhao2025ultrahr,
  title={UltraHR-100K: Enhancing UHR Image Synthesis with A Large-Scale High-Quality Dataset},
  author={Zhao, Chen and Ci, En and Xu, Yunzhe and Fan, Tiehan and Guan, Shanyan and Ge, Yanhao and Yang, Jian and Tai, Ying},
  journal={Advances in Neural Information Processing Systems},
  year={2025}
}

@article{zhao2024cycle,
  title={Cycle contrastive adversarial learning with structural consistency for unsupervised high-quality image deraining transformer},
  author={Zhao, Chen and Cai, Weiling and Hu, Chengwei and Yuan, Zheng},
  journal={Neural Networks},
  volume={178},
  pages={106428},
  year={2024},
  publisher={Elsevier}
}

@article{zhao2024learning,
  title={Learning a physical-aware diffusion model based on transformer for underwater image enhancement},
  author={Zhao, Chen and Dong, Chenyu and Cai, Weiling and Wang, Yueyue},
  journal={IEEE transactions on geoscience and remote sensing},
  year={2026}
}

@inproceedings{zhang2025gapt,
  title={GaPT-DAR: Category-level Garments Pose Tracking via Integrated 2D Deformation and 3D Reconstruction},
  author={Zhang, Li and Xu, Mingliang and Wang, Jianan and Yu, Qiaojun and Yang, Lixin and Li, Yonglu and Lu, Cewu and Wang, Rujing and Liu, Liu},
  booktitle={Proceedings of the Computer Vision and Pattern Recognition Conference},
  pages={22638--22647},
  year={2025}
}

@inproceedings{zhang2025r,
  title={R\^{} 2-Art: Category-Level Articulation Pose Estimation from Single RGB Image via Cascade Render Strategy},
  author={Zhang, Li and Jiang, Haonan and Huo, Yukang and Zhong, Yan and Wang, Jianan and Wang, Xue and Wang, Rujing and Liu, Liu},
  booktitle={Proceedings of the AAAI Conference on Artificial Intelligence},
  volume={39},
  number={9},
  pages={9985--9993},
  year={2025}
}

@article{zheng2023reparameterized,
  title={A reparameterized discrete diffusion model for text generation},
  author={Zheng, Lin and Yuan, Jianbo and Yu, Lei and Kong, Lingpeng},
  journal={arXiv preprint arXiv:2302.05737},
  year={2023}
}

@article{zhou2025canonical,
  title={Canonical Shape Reconstruction with SE (3) Equivariance Learning for Weakly-Supervised Object Pose Estimation},
  author={Zhou, Jun and Chen, Kai and Wei, Mingqiang and Zhang, Xiao-Ping and Dou, Qi and Qin, Jing},
  journal={IEEE Transactions on Circuits and Systems for Video Technology},
  year={2025},
  publisher={IEEE}
}

@inproceedings{karrer2011pinstripe,
  title={Pinstripe: eyes-free continuous input on interactive clothing},
  author={Karrer, Thorsten and Wittenhagen, Moritz and Lichtschlag, Leonhard and Heller, Florian and Borchers, Jan},
  booktitle={Proceedings of the SIGCHI Conference on Human Factors in Computing Systems},
  pages={1313--1322},
  year={2011}
}

@article{liu2021thin,
  title={Thin, soft, garment-integrated triboelectric nanogenerators for energy harvesting and human machine interfaces},
  author={Liu, Yiming and Yiu, Chunki and Jia, Huiling and Wong, Tszhung and Yao, Kuanming and Huang, Ya and Zhou, Jingkun and Huang, Xingcan and Zhao, Ling and Li, Dengfeng and others},
  journal={EcoMat},
  volume={3},
  number={4},
  pages={e12123},
  year={2021},
  publisher={Wiley Online Library}
}

@article{billard2019trends,
  title={Trends and challenges in robot manipulation},
  author={Billard, Aude and Kragic, Danica},
  journal={Science},
  volume={364},
  number={6446},
  pages={eaat8414},
  year={2019},
  publisher={American Association for the Advancement of Science}
}

@inproceedings{shridhar2022cliport,
  title={Cliport: What and where pathways for robotic manipulation},
  author={Shridhar, Mohit and Manuelli, Lucas and Fox, Dieter},
  booktitle={Conference on Robot Learning},
  pages={894--906},
  year={2022},
  organization={PMLR}
}

@article{amin2015comparative,
  title={Comparative study of augmented reality SDKs},
  author={Amin, Dhiraj and Govilkar, Sharvari},
  journal={International Journal on Computational Science \& Applications},
  volume={5},
  number={1},
  pages={11--26},
  year={2015},
  publisher={Academy and Industry Research Collaboration Center (AIRCC)}
}

@inproceedings{zhang2023genpose,
  title={GenPose: generative category-level object pose estimation via diffusion models},
  author={Zhang, Jiyao and Wu, Mingdong and Dong, Hao},
  booktitle={Proceedings of the 37th International Conference on Neural Information Processing Systems},
  pages={54627--54644},
  year={2023}
}

@inproceedings{zhang2024u,
  title={U-cope: Taking a further step to universal 9d category-level object pose estimation},
  author={Zhang, Li and Meng, Weiqing and Zhong, Yan and Kong, Bin and Xu, Mingliang and Du, Jianming and Wang, Xue and Wang, Rujing and Liu, Liu},
  booktitle={European Conference on Computer Vision},
  pages={254--270},
  year={2024},
  organization={Springer}
}

@inproceedings{rombach2022high,
  title={High-resolution image synthesis with latent diffusion models},
  author={Rombach, Robin and Blattmann, Andreas and Lorenz, Dominik and Esser, Patrick and Ommer, Bj{\"o}rn},
  booktitle={Proceedings of the IEEE/CVF conference on computer vision and pattern recognition},
  pages={10684--10695},
  year={2022}
}

@article{yang2023diffusion,
  title={Diffusion models: A comprehensive survey of methods and applications},
  author={Yang, Ling and Zhang, Zhilong and Song, Yang and Hong, Shenda and Xu, Runsheng and Zhao, Yue and Zhang, Wentao and Cui, Bin and Yang, Ming-Hsuan},
  journal={ACM Computing Surveys},
  volume={56},
  number={4},
  pages={1--39},
  year={2023},
  publisher={ACM New York, NY, USA}
}

@article{manas2023robotic,
  title={Robotic park: Multi-agent platform for teaching control and robotics},
  author={Ma{\~n}as-{\'A}lvarez, Francisco-Jos{\'e} and Guinaldo, Mar{\'\i}a and Dormido, Raquel and Dormido, Sebastian},
  journal={IEEE Access},
  volume={11},
  pages={34899--34911},
  year={2023},
  publisher={IEEE}
}

@article{jiang2023vima,
  title={Vima: Robot manipulation with multimodal prompts},
  author={Jiang, Yunfan and Gupta, Agrim and Zhang, Zichen and Wang, Guanzhi and Dou, Yongqiang and Chen, Yanjun and Fei-Fei, Li and Anandkumar, Anima and Zhu, Yuke and Fan, Linxi},
  year={2023}
}

@inproceedings{xue2023garmenttracking,
  title={Garmenttracking: Category-level garment pose tracking},
  author={Xue, Han and Xu, Wenqiang and Zhang, Jieyi and Tang, Tutian and Li, Yutong and Du, Wenxin and Ye, Ruolin and Lu, Cewu},
  booktitle={Proceedings of the IEEE/CVF Conference on Computer Vision and Pattern Recognition},
  pages={21233--21242},
  year={2023}
}

@article{myers1998brief,
  title={A brief history of human-computer interaction technology},
  author={Myers, Brad A},
  journal={interactions},
  volume={5},
  number={2},
  pages={44--54},
  year={1998},
  publisher={ACM New York, NY, USA}
}

@book{preece1994human,
  title={Human-computer interaction},
  author={Preece, Jenny and Rogers, Yvonne and Sharp, Helen and Benyon, David and Holland, Simon and Carey, Tom},
  year={1994},
  publisher={Addison-Wesley Longman Ltd.}
}

@inproceedings{chi2021garmentnets,
  title={Garmentnets: Category-level pose estimation for garments via canonical space shape completion},
  author={Chi, Cheng and Song, Shuran},
  booktitle={Proceedings of the IEEE/CVF International Conference on Computer Vision},
  pages={3324--3333},
  year={2021}
}

@article{austin2021D3PM,
  title={Structured denoising diffusion models in discrete state-spaces},
  author={Austin, Jacob and Johnson, Daniel D and Ho, Jonathan and Tarlow, Daniel and Van Den Berg, Rianne},
  journal={Advances in Neural Information Processing Systems},
  volume={34},
  pages={17981--17993},
  year={2021}
}

@article{hoogeboom2021argmax,
  title={Argmax flows and multinomial diffusion: Towards non-autoregressive language models},
  author={Hoogeboom, Emiel and Nielsen, Didrik and Jaini, Priyank and Forr{\'e}, Patrick and Welling, Max},
  journal={arXiv preprint arXiv:2102.05379},
  volume={3},
  number={4},
  pages={5},
  year={2021}
}

@inproceedings{sohl2015deep,
  title={Deep unsupervised learning using nonequilibrium thermodynamics},
  author={Sohl-Dickstein, Jascha and Weiss, Eric and Maheswaranathan, Niru and Ganguli, Surya},
  booktitle={International conference on machine learning},
  pages={2256--2265},
  year={2015},
  organization={PMLR}
}

@inproceedings{chen2020learning,
  title={Learning canonical shape space for category-level 6d object pose and size estimation},
  author={Chen, Dengsheng and Li, Jun and Wang, Zheng and Xu, Kai},
  booktitle={Proceedings of the IEEE/CVF conference on computer vision and pattern recognition},
  pages={11973--11982},
  year={2020}
}

@article{umeyama1991least,
  title={Least-squares estimation of transformation parameters between two point patterns},
  author={Umeyama, Shinji},
  journal={IEEE Transactions on Pattern Analysis \& Machine Intelligence},
  volume={13},
  number={04},
  pages={376--380},
  year={1991},
  publisher={IEEE Computer Society}
}

@article{yu2023comprehensive,
  title={A Comprehensive Survey of 3D Dense Captioning: Localizing and Describing Objects in 3D Scenes},
  author={Yu, Ting and Lin, Xiaojun and Wang, Shuhui and Sheng, Weiguo and Huang, Qingming and Yu, Jun},
  journal={IEEE Transactions on Circuits and Systems for Video Technology},
  year={2023},
  publisher={IEEE}
}

@article{carmigniani2011augmented,
  title={Augmented reality: an overview},
  author={Carmigniani, Julie and Furht, Borko},
  journal={Handbook of augmented reality},
  pages={3--46},
  year={2011},
  publisher={Springer}
}

@inproceedings{hou2021exploring,
  title={Exploring data-efficient 3d scene understanding with contrastive scene contexts},
  author={Hou, Ji and Graham, Benjamin and Nie{\ss}ner, Matthias and Xie, Saining},
  booktitle={Proceedings of the IEEE/CVF Conference on Computer Vision and Pattern Recognition},
  pages={15587--15597},
  year={2021}
}

@inproceedings{jaritz2019multi,
  title={Multi-view pointnet for 3d scene understanding},
  author={Jaritz, Maximilian and Gu, Jiayuan and Su, Hao},
  booktitle={Proceedings of the IEEE/CVF International Conference on Computer Vision Workshops},
  pages={0--0},
  year={2019}
}

@book{mason2001mechanics,
  title={Mechanics of robotic manipulation},
  author={Mason, Matthew T},
  year={2001},
  publisher={MIT press}
}

@article{couairon2022diffedit,
  title={Diffedit: Diffusion-based semantic image editing with mask guidance},
  author={Couairon, Guillaume and Verbeek, Jakob and Schwenk, Holger and Cord, Matthieu},
  journal={arXiv preprint arXiv:2210.11427},
  year={2022}
}

@article{alexanderson2023listen,
  title={Listen, denoise, action! audio-driven motion synthesis with diffusion models},
  author={Alexanderson, Simon and Nagy, Rajmund and Beskow, Jonas and Henter, Gustav Eje},
  journal={ACM Transactions on Graphics (TOG)},
  volume={42},
  number={4},
  pages={1--20},
  year={2023},
  publisher={ACM New York, NY, USA}
}

@article{liu2023audioldm,
  title={Audioldm: Text-to-audio generation with latent diffusion models},
  author={Liu, Haohe and Chen, Zehua and Yuan, Yi and Mei, Xinhao and Liu, Xubo and Mandic, Danilo and Wang, Wenwu and Plumbley, Mark D},
  journal={arXiv preprint arXiv:2301.12503},
  year={2023}
}

@inproceedings{ruiz2023dreambooth,
  title={Dreambooth: Fine tuning text-to-image diffusion models for subject-driven generation},
  author={Ruiz, Nataniel and Li, Yuanzhen and Jampani, Varun and Pritch, Yael and Rubinstein, Michael and Aberman, Kfir},
  booktitle={Proceedings of the IEEE/CVF conference on computer vision and pattern recognition},
  pages={22500--22510},
  year={2023}
}

@inproceedings{gu2022vector,
  title={Vector quantized diffusion model for text-to-image synthesis},
  author={Gu, Shuyang and Chen, Dong and Bao, Jianmin and Wen, Fang and Zhang, Bo and Chen, Dongdong and Yuan, Lu and Guo, Baining},
  booktitle={Proceedings of the IEEE/CVF conference on computer vision and pattern recognition},
  pages={10696--10706},
  year={2022}
}

@inproceedings{kawar2023imagic,
  title={Imagic: Text-based real image editing with diffusion models},
  author={Kawar, Bahjat and Zada, Shiran and Lang, Oran and Tov, Omer and Chang, Huiwen and Dekel, Tali and Mosseri, Inbar and Irani, Michal},
  booktitle={Proceedings of the IEEE/CVF Conference on Computer Vision and Pattern Recognition},
  pages={6007--6017},
  year={2023}
}

@article{ho2022cascaded,
  title={Cascaded diffusion models for high fidelity image generation},
  author={Ho, Jonathan and Saharia, Chitwan and Chan, William and Fleet, David J and Norouzi, Mohammad and Salimans, Tim},
  journal={Journal of Machine Learning Research},
  volume={23},
  number={47},
  pages={1--33},
  year={2022}
}

@article{xue2021omad,
  title={OMAD: Object Model with Articulated Deformations for Pose Estimation and Retrieval},
  author={Xue, Han and Liu, Liu and Xu, Wenqiang and Fu, Haoyuan and Lu, Cewu},
  journal={arXiv preprint arXiv:2112.07334},
  year={2021}
}

@article{liu2022toward,
  title={Toward real-world category-level articulation pose estimation},
  author={Liu, Liu and Xue, Han and Xu, Wenqiang and Fu, Haoyuan and Lu, Cewu},
  journal={IEEE Transactions on Image Processing},
  volume={31},
  pages={1072--1083},
  year={2022},
  publisher={IEEE}
}

@inproceedings{li2020category,
  title={Category-level articulated object pose estimation},
  author={Li, Xiaolong and Wang, He and Yi, Li and Guibas, Leonidas J and Abbott, A Lynn and Song, Shuran},
  booktitle={Proceedings of the IEEE/CVF Conference on Computer Vision and Pattern Recognition},
  pages={3706--3715},
  year={2020}
}

@inproceedings{di2022gpv,
  title={Gpv-pose: Category-level object pose estimation via geometry-guided point-wise voting},
  author={Di, Yan and Zhang, Ruida and Lou, Zhiqiang and Manhardt, Fabian and Ji, Xiangyang and Navab, Nassir and Tombari, Federico},
  booktitle={Proceedings of the IEEE/CVF Conference on Computer Vision and Pattern Recognition},
  pages={6781--6791},
  year={2022}
}

@inproceedings{che2024op,
  title={OP-Align: Object-level and Part-level Alignment for Self-supervised Category-level Articulated Object Pose Estimation},
  author={Che, Yuchen and Furukawa, Ryo and Kanezaki, Asako},
  booktitle={European Conference on Computer Vision},
  pages={72--88},
  year={2024},
  organization={Springer}
}

@article{wang2024dipose,
  title={DiPose: Discrete Diffusion Model for Occluded 3D Human Pose Estimation},
  author={Wang, Weiquan and Xiao, Jun and Wang, Chunping and Liu, Wei and Wang, Zhao and Chen, Long},
  journal={CoRR},
  year={2024}
}

@article{yang2023diffsound,
  title={Diffsound: Discrete diffusion model for text-to-sound generation},
  author={Yang, Dongchao and Yu, Jianwei and Wang, Helin and Wang, Wen and Weng, Chao and Zou, Yuexian and Yu, Dong},
  journal={IEEE/ACM Transactions on Audio, Speech, and Language Processing},
  volume={31},
  pages={1720--1733},
  year={2023},
  publisher={IEEE}
}

@article{zou20226d,
  title={6d-vit: Category-level 6d object pose estimation via transformer-based instance representation learning},
  author={Zou, Lu and Huang, Zhangjin and Gu, Naijie and Wang, Guoping},
  journal={IEEE Transactions on Image Processing},
  volume={31},
  pages={6907--6921},
  year={2022},
  publisher={IEEE}
}

@inproceedings{holmquist2023diffpose,
  title={Diffpose: Multi-hypothesis human pose estimation using diffusion models},
  author={Holmquist, Karl and Wandt, Bastian},
  booktitle={Proceedings of the IEEE/CVF international conference on computer vision},
  pages={15977--15987},
  year={2023}
}

@inproceedings{wang2023posediffusion,
  title={Posediffusion: Solving pose estimation via diffusion-aided bundle adjustment},
  author={Wang, Jianyuan and Rupprecht, Christian and Novotny, David},
  booktitle={Proceedings of the IEEE/CVF International Conference on Computer Vision},
  pages={9773--9783},
  year={2023}
}

@inproceedings{doosti2020hope,
  title={Hope-net: A graph-based model for hand-object pose estimation},
  author={Doosti, Bardia and Naha, Shujon and Mirbagheri, Majid and Crandall, David J},
  booktitle={Proceedings of the IEEE/CVF conference on computer vision and pattern recognition},
  pages={6608--6617},
  year={2020}
}

@article{chen2022analog,
  title={Analog bits: Generating discrete data using diffusion models with self-conditioning},
  author={Chen, Ting and Zhang, Ruixiang and Hinton, Geoffrey},
  journal={arXiv preprint arXiv:2208.04202},
  year={2022}
}

@inproceedings{wang2019normalized,
  title={Normalized object coordinate space for category-level 6d object pose and size estimation},
  author={Wang, He and Sridhar, Srinath and Huang, Jingwei and Valentin, Julien and Song, Shuran and Guibas, Leonidas J},
  booktitle={Proceedings of the IEEE/CVF Conference on Computer Vision and Pattern Recognition},
  pages={2642--2651},
  year={2019}
}

@inproceedings{liu2024kpa,
  title={KPA-Tracker: Towards Robust and Real-Time Category-Level Articulated Object 6D Pose Tracking},
  author={Liu, Liu and Huang, Anran and Wu, Qi and Guo, Dan and Yang, Xun and Wang, Meng},
  booktitle={Proceedings of the AAAI Conference on Artificial Intelligence},
  volume={38},
  number={4},
  pages={3684--3692},
  year={2024}
}

@inproceedings{liu2022akb,
  title={AKB-48: a real-world articulated object knowledge base},
  author={Liu, Liu and Xu, Wenqiang and Fu, Haoyuan and Qian, Sucheng and Yu, Qiaojun and Han, Yang and Lu, Cewu},
  booktitle={Proceedings of the IEEE/CVF Conference on Computer Vision and Pattern Recognition},
  pages={14809--14818},
  year={2022}
}

@inproceedings{katz2008manipulating,
  title={Manipulating articulated objects with interactive perception},
  author={Katz, Dov and Brock, Oliver},
  booktitle={2008 IEEE International Conference on Robotics and Automation},
  pages={272--277},
  year={2008},
  organization={IEEE}
}
}

\end{document}